\documentclass{article}
\usepackage{spconf,amsmath,graphicx,amssymb,algorithm,algorithmic}
\usepackage{color}

\title{BIAS MITIGATION POST-PROCESSING FOR INDIVIDUAL AND GROUP FAIRNESS}
\name{\begin{tabular}{c}Pranay K. Lohia,$^{\star,1}$ Karthikeyan Natesan Ramamurthy,$^{\star,2}$ Manish Bhide,$^{\dagger,3}$ \\Diptikalyan Saha,$^{\star,1}$ Kush R. Varshney,$^{\star,2}$ and Ruchir Puri$^{\dagger,2}$\end{tabular}}
\address{$^\star$IBM Research and $^\dagger$IBM Watson AI Platform \\$^1$Outer Ring Road, Embassy Manyatha B, Rachenahalli \& Nagawara Villages, Bangalore, KA, India \\$^2$1101 Kitchawan Road, Yorktown Heights, NY, USA \\$^3$Mindspace 3A, Hi-Tech City, Madhapur, Hyderabad, TG, India}
\begin{document}
\ninept
\maketitle
\begin{abstract}
Whereas previous post-processing approaches for increasing the fairness of predictions of biased classifiers address only group fairness, we propose a method for increasing both individual and group fairness.  Our novel framework includes an individual bias detector used to prioritize data samples in a bias mitigation algorithm aiming to improve the group fairness measure of disparate impact. We show superior performance to previous work in the combination of classification accuracy, individual fairness and group fairness on several real-world datasets in applications such as credit, employment, and criminal justice.
\end{abstract}
\section{Introduction}
\label{sec:intro}

Fairness, non-discrimination, and unwanted bias have always been concerns in human decision making \cite{VarshneyV2017}, but are increasingly in the limelight because historical human decisions are now being used as training data for machine learning models in high stakes applications such as employment, credit, and criminal justice \cite{WilliamsBS2018}.  Without bias mitigation, models trained on such decisions perpetuate and scale human biases and are thereby unsafe and untrustworthy \cite{VarshneyA2017,HindMMNROV2018}. The last couple of years have seen a surge in papers on algorithmic fairness in the machine learning and data mining literature, with basic principles defined using detection, estimation theory and information theory \cite{MenonW2018,CalmonWVRV2018}.

There are two main notions of fairness in decision making: \emph{group fairness} and \emph{individual fairness}.  Group fairness, in its broadest sense, partitions a population into groups defined by \emph{protected attributes} (such as gender, caste, or religion) and seeks for some statistical measure to be equal across groups. There are many different group fairness notions involving different statistical measures, one such notion being \emph{disparate impact} \cite{Narayanan2018}.  Individual fairness, in its broadest sense, seeks for similar individuals to be treated similarly. Checking for group fairness is a fairly straightforward computation of statistical metrics \cite{Zliobaitaz2017}, but checking for individual fairness is more computationally involved when there are many protected attributes with many values and scoring samples using a model is expensive \cite{GalhotraBM2017,AgarwalLNDS2018}. Unified metrics for both group and individual fairness have recently been proposed \cite{SpeicherHGGSWZ2018} based on inequality indices \cite{HurleyR2009}.  

Machine learning pipelines contain three possible points of intervention to mitigate unwanted bias: the training data, the learning procedure, and the output predictions, with three corresponding classes of bias mitigation algorithms: pre-processing, in-processing, and post-processing \cite{DalessandroOL2017}.  Advantages of post-processing algorithms are that they do not require access to the training process and are thus suitable for run-time environments. Moreover, post-processing algorithms operate in a black-box fashion, meaning that they do not need access to the internals of models, their derivatives, etc., and are therefore applicable to \emph{any} machine learning model (or amalgamation of models) \cite{KamiranKZ2012}.

The vast majority of bias mitigation algorithms address group fairness, but a few address individual fairness \cite{DworkHPRZ2012,DworkI2018}.  Some pre-processing algorithms address both group and individual fairness \cite{ZemelWSPD2013,CalmonWVRV2017,CalmonWVRV2018}, but to the best of our knowledge, all existing post-processing algorithms are only for group fairness \cite{KamiranKZ2012,HardtPS2016,PleissRWKW2017,CanettiCDRSS2018}. Our main contribution in this paper is to propose a post-processing bias mitigation algorithm that considers \emph{both} group and individual fairness. Moreover, unlike the previous work, our proposal does not require any ground truth class labels in the validation samples while training the bias mitigation algorithm.

The general methodology of post-processing algorithms is to take a subset of samples and change their predicted labels appropriately to meet a group fairness requirement.  An interesting observation about post-processing is that \emph{any} sample can be altered to achieve group fairness requirements because the metrics are expectations.  The papers \cite{HardtPS2016,PleissRWKW2017} choose the samples randomly, whereas \cite{KamiranKZ2012} chooses the most uncertain samples (the ones in the reject option band \cite{Chow1970,Varshney2011}), capturing the human intuition to give the benefit of the doubt to unprivileged groups.  In the method we propose herein, we choose samples that have or are likely to have individual fairness issues and in this way are able to address both group and individual fairness together.

The starting point for our proposed approach is the individual bias detector of  \cite{AgarwalLNDS2018}, which finds samples whose model prediction changes when the protected attributes change, leaving all other features constant.  Despite a large set of efficiencies enacted in the algorithm, it is still computationally expensive.  To overcome the limitation of not being able to run the detector continually, we check for individual fairness on a small set of points and generalize from them by training a classifier that is applied to new samples.  The samples with likely individual bias are the ones considered for a change of predicted label.  By doing so, we modify the idea of \cite{KamiranKZ2012} from focusing on uncertainty to focusing on individual bias.

Our empirical results are promising.  Compared to the state-of-the-art algorithms of \cite{HardtPS2016} and \cite{KamiranKZ2012}, we have superior performance on the combination of classification accuracy, individual fairness, and group fairness in the preponderance of six different real-world classification tasks requiring non-discrimination. The results show very little reduction in classification accuracy with much improvement in individual and group fairness measures. 

The remainder of the paper is organized as follows.  We first provide background on individual and group fairness definitions and detectors in Sec.~\ref{sec:background}.  Next, in Sec.~\ref{sec:algorithm}, we propose a new post-processing bias mitigation algorithm that accounts for both individual and group fairness.  In Sec.~\ref{sec:results}, we provide empirical results on several real-world datasets including comparisons to \cite{KamiranKZ2012,HardtPS2016}.  Finally, we conclude the paper in Sec.~\ref{sec:conclusion}.

\section{Individual and Group Fairness}
\label{sec:background}

In this section, we introduce notation, provide working definitions of individual and group fairness, and detail methods for detecting individual bias and mitigating group bias. 

Consider a supervised classification problem with features $\mathbf{X} \in \mathcal{X}$, categorical protected attributes $\mathbf{D} \in \mathcal{D}$, and categorical labels $Y \in \mathcal{Y}$.  We are given a set of training samples $\{(\mathbf{x}_1,\mathbf{d}_1,y_1), \ldots, (\mathbf{x}_n,\mathbf{d}_n,y_n)\}$ and would like to learn a classifier $\hat{y}: \mathcal{X} \times \mathcal{D} \rightarrow \mathcal{Y}$. For ease of exposition, we will only consider a scalar binary protected attribute, i.e.\ $\mathcal{D} = \{0,1\}$, and a binary classification problem, i.e.\ $\mathcal{Y} = \{0,1\}$.\footnote{In many realistic settings, these simplifications do not hold, which motivate the individual bias detector component described in Sec.~\ref{sec:algorithm:ind}.}  The value $d = 1$ is set to correspond to the \emph{privileged} group (e.g.\ whites in the United States in criminal justice applications) and $d = 0$ to \emph{unprivileged} group (e.g.\ blacks).  The value $y = 1$ is set to correspond to a \emph{favorable} outcome (e.g.\ receiving a loan or not being arrested) and $y = 0$ to an \emph{unfavorable} outcome. Based on the context, we may also deal with probabilistic binary classifiers with continuous output scores $\hat{y}_S \in [0,1]$ that are thresholded to $\{0,1\}$.

One definition of individual bias is as follows.  Sample $i$ has individual bias if $\hat{y}(\mathbf{x}_i,d=0) \neq \hat{y}(\mathbf{x}_i,d=1)$. Let $b_i = I[\hat{y}(\mathbf{x}_i,d=0) \neq \hat{y}(\mathbf{x}_i,d=1)]$, where $I[\cdot]$ is an indicator function. The individual bias score, $b_{S, i} = \hat{y}_S(\mathbf{x}_i,d=1) - \hat{y}_S(\mathbf{x}_i,d=0)$, is a soft version of $b_i$. To compute an individual bias summary statistic, we take the average of $b_i$ across test samples.

One notion of group fairness known as \emph{disparate impact} is defined as follows.  There is disparate impact if
\begin{equation}
    \label{eqn:disp_imp}
    \frac{\mathbb{E}[\hat{y}(\mathbf{X},D) \mid D = 0]}{\mathbb{E}[\hat{y}(\mathbf{X},D) \mid D = 1]}
\end{equation}
is less than $1 - \epsilon$ or greater than $(1 - \epsilon)^{-1}$, where a common value of $\epsilon$ is 0.2.

\subsection{Test Generation for Individual Bias Detection}

There are two distinct problems in individual bias detection: first, determining whether there are any cases of individual bias, and second, determining the individual bias status of all samples.  
In our earlier work~\cite{AgarwalLNDS2018}, we presented a technique for the first problem that systematically explores the decision space of any black box classifier to generate test samples that have an enhanced chance of being biased.  The method uses two kinds of search: (a) a global search which explores the decision space such that diverse areas are covered, and (b) a local search which generates test cases by intelligently perturbing the values of non-protected features of an already found individually-biased sample. The key idea is to use dynamic symbolic execution, an existing systematic test case generation technique for programs that generates search constraints by negating the constraints in a program path and uses a constraint solver to find new search paths \cite{DART}.  
This algorithm is useful in solving the second of the distinct problems from a computational perspective when used on a batch of samples in settings involving a large number of attributes and attribute values.

\subsection{Post-Processing to Achieve Group Fairness}

To achieve acceptable group fairness, various post-processing methods may be applied to change the label outputs of the classifier $\hat{y}_i$ to other labels $\check{y}_i \in \mathcal{Y}$.  The reject option classification (ROC) method of \cite{KamiranKZ2012} considers \emph{uncertain} samples with $|\hat{y}_S-0.5| < \theta$ (assuming $0.5$ is the classification threshold) for some margin parameter $\theta$ and assigns $\check{y}_i = 1$ for samples with $d_i = 0$ and assigns $\check{y}_i = 0$ for samples with $d_i = 1$.  For \emph{certain} samples outside the so-called reject option band, $\check{y}_i = \hat{y}_i$.  The $\theta$ value may be optimized to achieve the requirement on disparate impact.

The algorithm proposed by \cite{HardtPS2016}, equalized odds post-processing (EOP), is targeted to a different group fairness measure: equalized odds rather than disparate impact.  Perfect equalized odds requires the privileged and unprivileged groups to have the same false negative rate and same false positive rate.  The algorithm solves an optimization problem to find probabilities with which to assign $\check{y}_1 = 1 - \hat{y}_i$.  There are four such probabilities for the following four combinations: $(d_i = 0, \hat{y} = 0)$, $(d_i = 0, \hat{y} = 1)$, $(d_i = 1, \hat{y} = 0)$, and $(d_i = 1, \hat{y} = 1)$.  With these probabilities, the individual points whose prediction is flipped is a random draw.  The methods of \cite{PleissRWKW2017,CanettiCDRSS2018} are refinements of \cite{HardtPS2016} and share the key characteristics. 

\section{Proposed Algorithm}
\label{sec:algorithm}

The new fairness post-processing algorithm we propose is inspired by and not radically different from \cite{KamiranKZ2012} in form.  The key observation in post-processing for group fairness metrics like disparate impact is that since they are defined as expectations, the individual samples are exchangeable. Kamiran et al.~\cite{KamiranKZ2012} elect to change values of $\hat{y}_i$ to $\check{y}_i$ in a reject option band to conform to one type of human sensibility, but the same effect on disparate impact can be achieved using the same numbers of samples from elsewhere in $\mathcal{X}$.  And that is precisely what we propose: elect samples from parts of $\mathcal{X}$ that likely have individual bias.  In this section, we first describe individual bias detection and then how we wrap that in a post-processing bias mitigation algorithm.

\subsection{Individual Bias Detector}
\label{sec:algorithm:ind}

Consider a classifier $\hat{y}$ already trained on a training dataset partition. We can evaluate the individual bias definition provided in Sec.~\ref{sec:background} on a validation partition that has no labels to go alongside.  Some of these validation samples will have individual bias and some will not.  Under an assumption of some coherence or smoothness of individual bias in $\mathcal{X}$, we can learn a classifier or detector for individual bias from this validation set that will generalize to unseen samples whose individual bias is unknown.  One may use any classification or anomaly detection algorithm here that provides score outputs.  We use logistic regression in the empirical results.

Formally, by perturbing the $d_j$ of validation set samples $(\mathbf{x}_j,d_j)$, $j = 1,\ldots,m$, that belong to the unprivileged group ($d_j = 0$), we obtain individual bias scores $b_{S, j}$. We construct a further dataset $\{(\mathbf{x}_1,\beta_1),\ldots,(\mathbf{x}_m,\beta_m)\}$, and use it to train an individual bias detector $\hat{b}(\cdot)$.
$\beta_j$ is 1 for the samples that have the highest individual bias scores, and 0 for the rest. This assignment is determined by a threshold $\tau$ on the individual bias scores chosen based on the disparate impact constraint on the entire validation set. This is similar to the ROC algorithm where the margin parameter is adjusted based on disparate impact requirements.

One may argue that a trained individual bias detector is unnecessary and one should simply compute $b_i$ for all samples as they come at run-time because doing so only involves scoring using the black-box classifier model.  This may be true, but with the following caveats.  Firstly, in the exposition of the paper, we have assumed $d_i$ to be scalar and binary, when in many instances it is not.  Therefore, computing $b_i$ may require several model evaluations which could be prohibitive, especially in the industrial usage we imagine in which each sample that is scored costs a certain amount of money to be paid by the entity deploying the model and remediating the bias.  Secondly, we compute the binary $\beta_j$ values based on the group fairness constraint, which ensures that only examples with highest individual bias scores are considered for debiasing, and there is no overcompensation. This level of control is not possible if we consider all examples with $b_i=1$ to be equally biased. 


\subsection{Overall Algorithm}

Once we have the individual bias detector $\hat{b}$ 
trained on the validation set, the bias mitigation algorithm applied in run-time to test samples is as follows.  Each sample from the unprivileged group ($d_i = 0$) is tested for individual bias and if it is likely to have individual bias, i.e., $\hat{b}_i = 1$, then this sample is assigned the outcome it would have received if it were in the privileged group, i.e., $\check{y}_i = \hat{y}(\mathbf{x}_k,1)$.  To encode a human sensibility similar to ROC, all other samples are left unchanged, including samples from the privileged group. 


The proposed algorithm is summarized below:
\begin{algorithm}
  \caption{Individual+Group Debiasing (IGD) Post-Processing}
  \label{algo:igd}
  \begin{algorithmic}[t]
  \STATE{Given classifier $\hat{y}$ trained on training set $\{(\mathbf{x}_i,d_i,y_i)\}$, and}
  \STATE{Given validation set $\{\mathbf{x}_j \mid d_j = 0\}$, compute individual bias scores $\{b_{S, j} \mid d_j = 0 \}$.}
  \IF{$b_{S, j} > \tau$}
    \STATE{$\beta_j \leftarrow 1$}
  \ELSE
    \STATE{$\beta_j \leftarrow 0$}
  \ENDIF
  \STATE{Construct auxiliary dataset $\{(\mathbf{x}_j,\beta_j) \mid d_j = 0 \}$.}
  \STATE{Train individual bias detector $\hat{b}$ on auxiliary dataset.}
  \FORALL{run-time test samples $(\mathbf{x}_k,d_k)$}
    \STATE{$\hat{y}_{k} \leftarrow \hat{y}(\mathbf{x}_k,d_k)$}
    \IF{$d_{k} == 0$}
        \STATE{$\hat{b}_k \leftarrow \hat{b}(\mathbf{x}_k)$}
        \IF{$\hat{b}_k == 1$} 
            \STATE{$\check{y}_k \leftarrow \hat{y}(\mathbf{x}_k,1)$}
        \ELSE
            \STATE{$\check{y}_k \leftarrow \hat{y}_k$}
        \ENDIF
    \ELSE
    	\STATE{$\check{y}_k \leftarrow \hat{y}_k$}
    \ENDIF
  \ENDFOR
  \end{algorithmic}
 \end{algorithm}

\section{Empirical Results}
\label{sec:results}

We evaluate our proposed algorithm on three standard datasets: UCI Adult (an income dataset based on a 1994 US Census database; 45,222 samples; favorable outcome: income greater than \$50,000; protected attributes: sex, race), UCI Statlog German Credit (a credit scoring dataset; 1,000 samples; favorable outcome: low risk; protected attributes: sex, age), and ProPublica COMPAS (a prison recidivism dataset; 6,167 samples. favorable outcome: does not reoffend; protected attributes: sex, race). Each of the three datasets has two binary protected attributes that we consider as two different problems, yielding six problems overall. 
We compare our proposed individual+group debiasing (IGD) algorithm with ROC \cite{KamiranKZ2012} and EOP \cite{HardtPS2016} using the implementations of ROC and EOP provided in the AI Fairness 360 toolkit \cite{aif360}.

We process and load each dataset using the AI Fairness 360 toolkit and randomly divide it into 60\% training, 20\% validation and 20\% testing partitions. We conduct experiments with 25 such random partitions of the datasets, allowing us to provide error bars in the empirical results that follow. Using the training partition, we fit both $\ell_2$-regularized logistic regression and random forests as black-box classifiers. For random forests, we set the number of trees to 100 and the minimum samples per leaf node to 20. 

 The parameters of all three bias mitigation approaches are optimized on the validation partition of the dataset. Both the ROC and the EOP approaches require ground truth class labels in the validation set, whereas the proposed IGD approach, being a pure run-time method, does not. ROC and IGD are optimized to achieve disparate impact in the range $(0.8,1.25)$, i.e., $\epsilon = 0.2$. EOP, being designed for equalized odds rather than disparate impact,  cannot be optimized for ranges of disparate impact. 

In the subsections that follow, we first demonstrate the efficacy of the individual bias detector used in the proposed IGD algorithm and then compare the three algorithms for classification accuracy, disparate impact, and individual fairness.

\subsection{Validation Results on Individual Bias Generalization}
\label{sec:ind_bias_gen}
We verify the generalization performance of the individual bias detector on unseen test data. Since the individual bias detector is used only on unprivileged group samples ($d = 0$), its performance measure is only computed for this subset. The ground truth labels for the bias detector are obtained by actually computing the individual bias scores ($b_{S,k}$) for all unprivileged group samples in the test data, and identifying the ground truth bias labels ($\beta_k$) based on the disparate impact constraint. These labels are compared with the labels predicted by the bias detector ($\hat{b}_k$), and the balanced classification accuracy is computed. 

\begin{figure}
  \centering
  \includegraphics[width=3.2in]{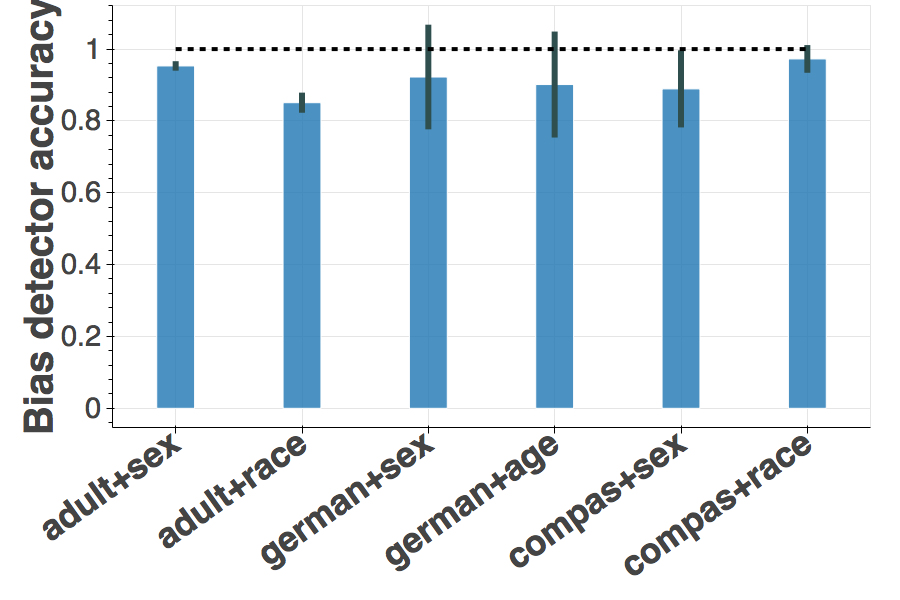}
\caption{Balanced accuracy of the bias detector when the black box classifier is a Logistic Regression model. The bar shows the mean accuracy, and the vertical lines show the extent of $\pm 1$ standard deviation. The dotted horizontal line shows the best possible performance.}
\label{fig:bias_det_acc_lr}
\end{figure}

This performance of the bias detector is shown in Fig.~\ref{fig:bias_det_acc_lr} for all dataset and protected attribute combinations when the black-box classifier is logistic regression. All accuracy values are more than 0.85, which illustrates its clear effectiveness for the purpose at hand. The detector performs similarly when the black-box classifier is random forests, with a minimum accuracy of approximately 0.80. 

\subsection{Fairness Comparisons}
\label{sec:fairness_comp}
\begin{figure}[ht]
  \centering
  \includegraphics[width=3.2in]{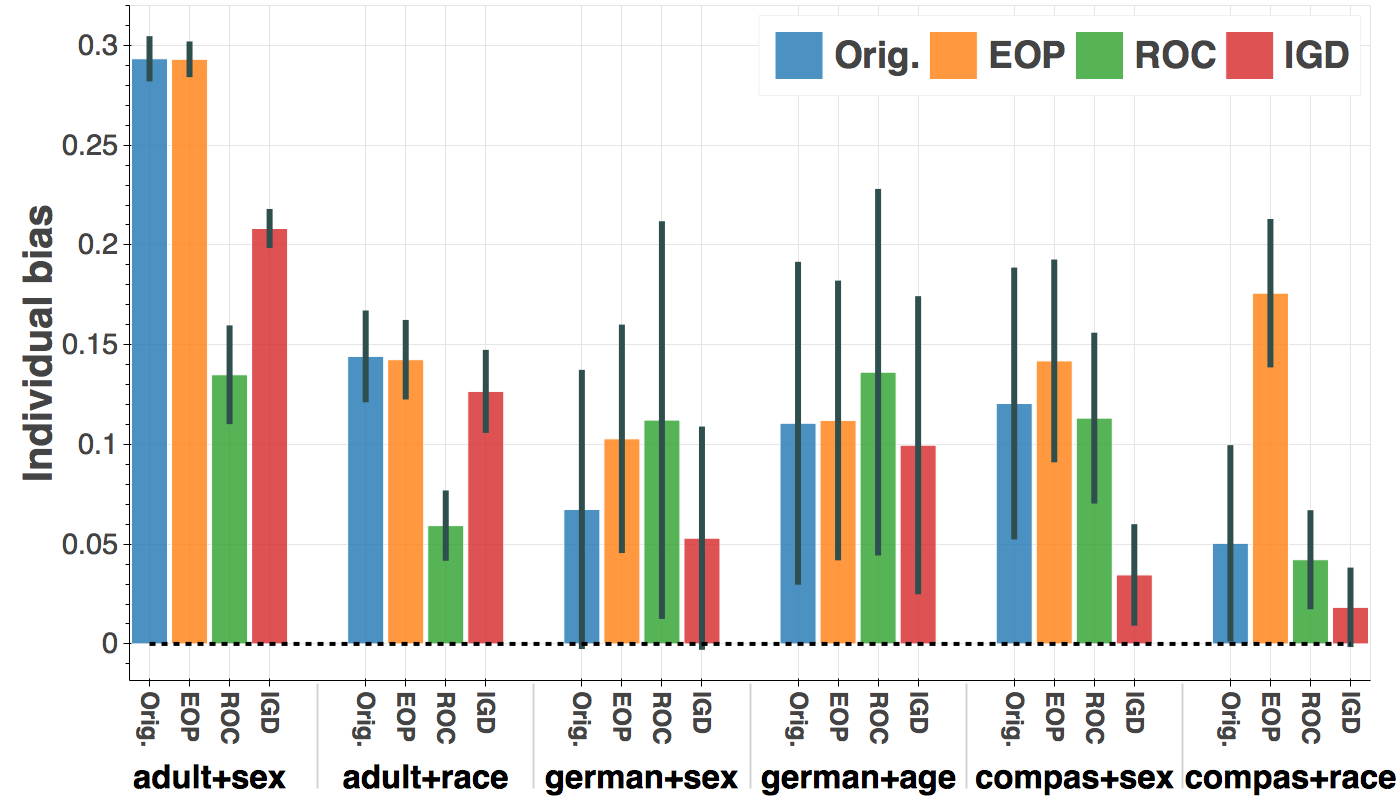}
\caption{Individual bias of the original model and the compared post-processing algorithms. The bar shows the mean value, and the vertical lines show the extent of $\pm 1$ standard deviation. The dotted horizontal line shows the ideal fair value (0.0).}
\label{fig:lr_ind_bias}
\end{figure}

\begin{figure}[ht]
  \centering
  \includegraphics[width=3.2in]{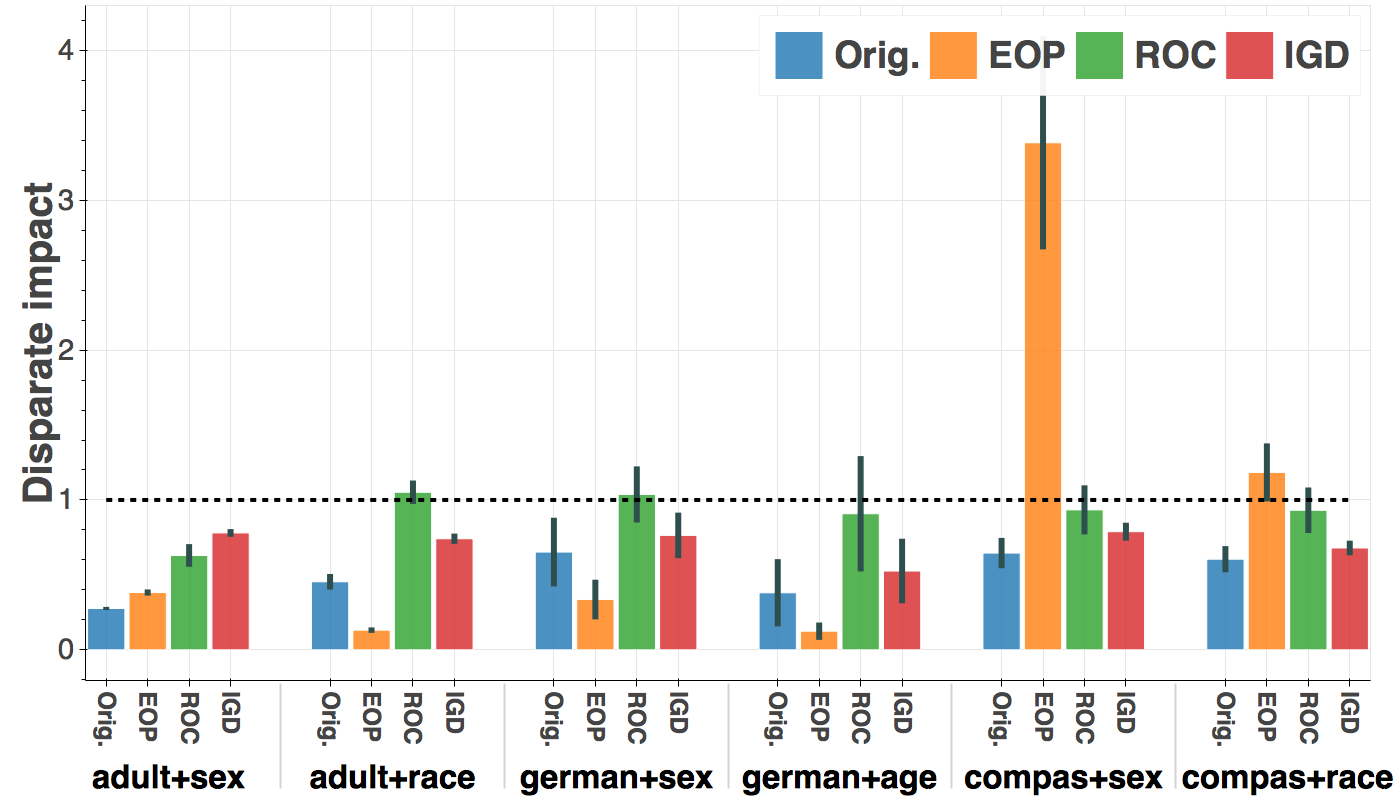}
\caption{Disparate impact of the original model and the compared post-processing algorithms. The bar shows the mean value, and the vertical lines show the extent of $\pm 1$ standard deviation. The dotted horizontal line shows the ideal fair value (1.0).}
\label{fig:lr_disp_imp}
\end{figure}

\begin{figure}[ht]
  \centering
  \includegraphics[width=3.2in]{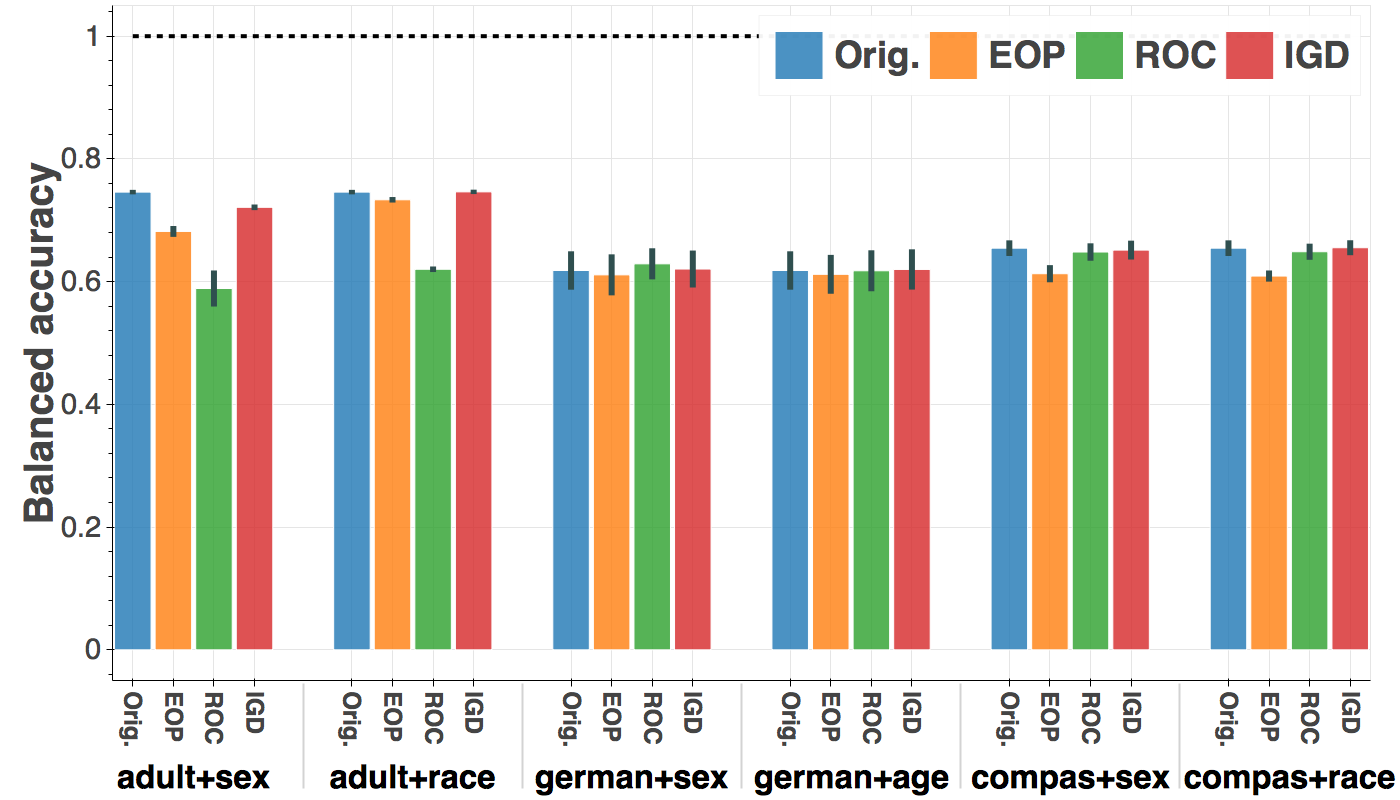}
\caption{Balanced classification accuracy of the original model and the compared post-processing algorithms. The bar shows the mean value, and the vertical lines show the extent of $\pm 1$ standard deviation. The dotted horizontal line is the best possible accuracy (1.0).}
\label{fig:lr_bal_acc}
\end{figure}

We use three measures for comparing EOP, ROC, and IGD: (a) individual bias, (b) disparate impact, and (c) balanced classification accuracy. These measures are computed using the post-processed predictions $\check{y}$. The individual bias measure is the summary statistic discussed in Sec. \ref{sec:background}, the disparate impact measure is defined in (\ref{eqn:disp_imp}), and balanced classification accuracy is the mean of true positive and true negative rates obtained for the predictions $\check{y}$ with respect to the true labels $y$. We also obtain these measures for the  original (Orig.) predictions $\hat{y}$. As shown in Fig.~\ref{fig:lr_ind_bias}, Fig.~\ref{fig:lr_disp_imp}, and Fig.~\ref{fig:lr_bal_acc}, the proposed IGD approach is the only one that consistently improves both fairness measures while keeping the accuracy close to that of the original classifier.  All results are show for logistic regression as the black-box classifier, but similar results are also observed for random forests (omitted due to space constraints).


In individual bias, the proposed IGD method performs the best for the German and COMPAS datasets. The ROC method performs the best for the Adult dataset, at the expense of reducing the balanced accuracy. Sometimes the EOP and ROC methods increase the individual bias, which is never the case with IGD. The proposed IGD method also consistently improves disparate impact over the original predictions, although outperformed by the ROC method in five out of six cases. The strong performance of the ROC approach is likely because it does not also optimize for individual bias. The EOP method performs poorly on disparate impact, likely because it was designed to equalize odds, which may or may not always result in improved disparate impact \cite{FriedlerSVCHR2018}. The proposed IGD method is also the best in preserving the balanced classifier accuracy compared to the original predictions even though no ground truth labels are used in the validation partition.



\section{Conclusion}
\label{sec:conclusion}

Algorithmic fairness is an important topic for business and society, and developing new bias mitigation algorithms that address as many facets of fairness as possible is critical.  In this paper, we have developed a new post-processing algorithm that targets samples with individual bias for remediation in order to improve \emph{both} individual and group fairness metrics and shown that it does so empirically on several real-world datasets without much loss in classification accuracy.  From our experience, the machine learning industry is moving towards paradigms in which there will be a separation between model building and model deployment. This will include a limited ability for deployers to gain access to the internals of pre-trained models. Therefore, post-processing algorithms, especially ones that can treat a classifier as a complete black-box are necessary.  In comparison to previous work, our proposed algorithm not only tackles both individual and group fairness, but also is a pure run-time approach because it does not require ground truth class labels for the validation set.

\bibliographystyle{IEEEtran}
\bibliography{postprocessing}

\end{document}